\documentclass{article}

\usepackage[preprint,nonatbib]{neurips_2022}

\usepackage[utf8]{inputenc} 
\usepackage[T1]{fontenc}    
\usepackage{hyperref}       
\usepackage{url}            
\usepackage{booktabs}       
\usepackage{amsfonts}       
\usepackage{nicefrac}       
\usepackage{microtype}      
\usepackage{xcolor}         

\usepackage[
    natbib=true,
    citestyle=numeric-comp,
    bibstyle=ieee,
    sorting=nty
]{biblatex}
\usepackage{graphicx}
\usepackage{caption}
\usepackage{subcaption}
\usepackage{bm}
\usepackage{amsmath,amssymb}
\usepackage{booktabs}
\usepackage{algorithm}
\usepackage[noend]{algorithmic}
\usepackage{multirow}
\usepackage[flushleft]{threeparttable}
\usepackage{makecell}

\hypersetup{
    colorlinks=true,
    linkcolor=red,
    citecolor=cyan,
    filecolor=magenta,      
    urlcolor=cyan,
}

\definecolor{f3green}{RGB}{0,144,81}
\definecolor{f3brown}{RGB}{148,82,0}

\DeclareMathOperator*{\ce}{\texttt{cross\_entropy}}

\DeclareMathOperator*{\clstoken}{\texttt{[CLS]}}

\newcommand{\ie}{\textit{i}.\textit{e}.}
\newcommand{\eg}{\textit{e}.\textit{g}.}
\newcommand{\ours}{\texttt{DoPrompt}}

\usepackage[textsize=tiny]{todonotes}
\usepackage{hhline}
\usepackage{colortbl}
\definecolor{Gray}{gray}{0.9}
\definecolor{ForestGreen}{rgb}{0.13, 0.55, 0.13}
\definecolor{ForestGreen2}{rgb}{0.2, 0.5372549019607843, 0.1803921568627451}
\definecolor{newGREEN}{RGB}{34,139,34}
\definecolor{myMaroon}{RGB}{231, 52, 52}
\definecolor{Maroon}{rgb}{0.69, 0.19, 0.0}
\definecolor{my_cyan}{RGB}{112, 242, 244}
\newcommand{\FG}[1]{\textcolor{ForestGreen}{#1}}
\newcommand{\XY}[1]{\textcolor{red}{#1}}
\newcommand\Tstrut{\rule{0pt}{2.6ex}}         

\newcommand{\ulbf}[1]{\underline{\bf #1}}
\usepackage{wrapfig}

\title{Prompt Vision Transformer for Domain Generalization}

\author{Zangwei Zheng$^{1,}$ \quad Xiangyu Yue$^2$  \quad Wang Kai$^1$ \quad Yang You$^1$\\
$^1$National University of Singapore \quad
$^2$Berkeley\\
\{zangwei, kai.wang, youy\}@comp.nus.edu.sg \quad xyyue@berkeley.edu
}

\begin{document}

\maketitle

\begin{abstract}
  Though vision transformers (ViTs) have exhibited impressive ability for representation learning, we empirically find that they cannot generalize well to unseen domains with previous domain generalization algorithms. In this paper, we propose a novel approach~\ours~based on prompt learning to embed the knowledge of source domains in domain prompts for target domain prediction. Specifically, domain prompts are prepended before ViT input tokens from the corresponding source domain. Each domain prompt learns domain-specific knowledge efficiently since it is optimized only for one domain. Meanwhile, we train a prompt adapter to produce a suitable prompt for each input image based on the learned source domain prompts. At test time, the adapted prompt generated by the prompt adapter can exploit the similarity between the feature of the out-of-domain image and source domains to properly integrate the source domain knowledge.
  Extensive experiments are conducted on four benchmark datasets. Our approach achieves 1.4\% improvements in the averaged accuracy, which is 3.5$\times$ the improvement of the state-of-the-art algorithm with a ViT backbone.

\end{abstract}

\section{Introduction}

Deep learning has achieved remarkable performances in various computer vision tasks~\cite{He2016DeepRL, Redmon2016YouOL, Shelhamer2017FullyCN}. With the development of network backbones, the learned representation is more generalizable when evaluated in the i.i.d. setting (independent and identical distribution). However, the i.i.d. setting is often violated in the real world due to the remarkable domain shift between training and test data. To tackle this problem, domain generalization (DG) aims to learn a generalizable model from a set of source domains and perform well on the unseen target domain.


Many works in domain generalization propose to learn a domain-invariant feature across source domains \cite{ghifary2016scatter,motiian2017unified,li_domain_2018}. However, when source domains are very diverse, it is difficult to learn a domain-invariant model because each domain contains much domain-specific knowledge. Besides, domain-invariant features among source domains cannot ensure a small domain gap between the source and unseen target domains. Recent works~\cite{Ding2018DeepDG,Seo2020LearningTO,Zhou2021DomainAE} propose to give the prediction for an unseen target domain based on different experts in each source domain. Specifically, they divide network parameters into two sets. The domain-specific one contains components specific for each source domain, and the rest parameters are shared across all domains. In this way, the network can capture the specialized pattern in each domain more efficiently.

Domain-experts-based methods help the model to predict in an unseen test domain. However, two main challenges still remain in the previous methods. First, the specialized components should have the ability to capture different domain knowledge. Previous methods only choose some parameters (\eg, bias, shallow layers, batchnorm) of the network and cannot interact with the whole network. This limits the ability of the network to fully incorporate the domain knowledge. In addition, since there is no access to the target domain, we need a well-designed machanism to extend the learned domain-specific knowledge for the target image at the inference time. \cite{Ding2018DeepDG} uses an ensemble method by taking the average of output by the model with different source domain components and \cite{Zhou2021DomainAE} adopts the most confident prediction. Both are ignorant of the relationship and similarity between input target images and source domains, and thus cannot fully take advantage of different source knowledge.

To tackle the first problem, we exploit the prompt learning to build powerful domain experts. 
Vision transformers (ViTs) have achieved striking performance in computer vision tasks~\cite{Dosovitskiy2021AnII,Liu2021SwinTH}. 
With the self-attention module, ViTs can fully model the relationship between different image patches. Prompt learning~\cite{Zhou2021LearningTP,Liu2021PTuningVP} is a popular method for downstream task learning in NLP. For different downstream tasks, the knowledge about the task is embedded into the input tokens. Since tokens can interact with the prompt tokens at all layer levels through self-attention layers, the network can well understand the meaning of the task if the prompt carries enough information. Inspired by this, we hope to embed the different domain knowledge into different domain prompts. Compared with previous methods, the domain prompts are more powerful in carrying different knowledge, which means they can more efficiently capture the pattern in different domains.

For the second challenge, we propose to learn how to take better advantage of source information. Matching Network~\cite{vinyals2016matching} proposes to adaptively use the seen information for the unseen one in the one-shot learning. In the one-shot generalization problem, the source knowledge is semantic information used to assign the class label for the test image. In contrast, for domain generalization, the source knowledge is about the domain information, which should not affect the semantic of each image. Thus, the design of adative knowledge combination should be different from the attention mechanism proposed in~\cite{vinyals2016matching}. We propose a domain adapter that can analyze the domain distance between the input image and the source domains in the feature space and produces a better prompt for extracting a better representation of images in the unseen domain.

To enable more powerful domain experts and domain adapter, we propose~\ours, which consists of two objectives during training: Domain Prompt Learning (DPL) and Prompt Adapter Learning (PAL). For Domain Prompt Learning, image patch tokens are prepended with the prompt token of the corresponding domain. The domain prompt learns the domain-specific knowledge in each source domain. To take full advantage of the domain-specific knowledge for target domain prediction,
we use a prompt adapter to integrate the learned source domain prompts to generate an well-adapted prompt for each target image. In order to learn a good prompt adapter, we encourage it to distinguish between features from different domains and produce prompts contributive to a correct prediction. During inference time, the adapted prompt generated by the prompt adapter is used for out-of-distribution prediction. Experiments show that the adapted prompt can reflect similarities between input images and different source domains.

We evaluate previous domain generalization methods and our algorithm based on ViT architecture with four standard benchmarks: PACS~\cite{Li2017DeeperBA}, VLCS~\cite{Fang2013UnbiasedML}, OfficeHome~\cite{Venkateswara2017DeepHN}, and DomainNet~\cite{Peng2019MomentMF}. With a controlled computing condition similar to \cite{gulrajani_search_2020}, the proposed \ours~outperforms state-of-the-art methods with ResNet-50 by 4.7\% on the averaged accuracy of four datasets. \ours~outperforms the baseline method with ViT-base as the backbone by 1.4\%, which is 3.5$\times$ the best improvement that previous performance can achieve. 

To summarize, our contributions are as follows:
\begin{itemize}
 \item We benchmark previous DG 
 algorithms on ViTs and demonstrate that while ViTs outperform ResNet-50, many previous methods can hardly improve the performance of domain generalization with the ViT backbone.
 \item We propose an effective domain generalization algorithm \ours~for vision transformers with Domain Prompt Learning (DPL) and Prompt Adapter Learning (PAL).
 \item To the best of our knowledge, this is the first domain generalization algorithm considering the unique architecture of vision transformers, which outperforms state-of-the-art methods by a large
 margin across multiple benchmark datasets.
\end{itemize}

\section{Related Work}
\paragraph{Domain generalization with different architectures.} Many domain generalization approaches have been proposed during the past decade based on the strong ability of deep neural networks to learn a good representation. Early works~\cite{Bousmalis2016DomainSN,Li2017DeeperBA} on domain generalization evaluated their methods on AlexNet~\cite{Krizhevsky2012ImageNetCW}. In an attempt to show the approach can be applied to different network backbones, many works~\cite{li_learning_2017,Carlucci2019DomainGB} conducted experiments both on AlexNet and ResNet-18~\cite{He2016DeepRL}. However, the two networks are both convolutional neural networks and too shallow from nowadays view. Recently, an increasing number of works~\cite{Balaji2018MetaRegTD,Nam2021ReducingDG,Huang2020SelfChallengingIC} applied their methods with ResNet-50 pretrained on ImageNet~\cite{Deng2009ImageNetAL}. Among them, \cite{gulrajani_search_2020} benchmarked a large number of previous approaches on the ResNet-50 backbone with controlled computing budgets. It demonstrated that under the same setting, the basic Empirical Risk Minimization (ERM) loss outperforms many previous methods. This result shows that reconsidering the efficacy of DG methods with different backbones is necessary. In this paper, we investigate whether the past algorithms still work for the transformer-based vision model.

\paragraph{Vision transformer.} \cite{Dosovitskiy2021AnII} showed that pure transformer architecture could achieve state-of-the-art performance in image classification. Recently, \cite{zhang2022delving} and \cite{ge2022domain} propose to use ViTs for domain adaptation. Both of them are based on the aligment method and cannot be extended to the domain generalization setting. \cite{Raghu2021DoVT,Naseer2021IntriguingPO} found that ViTs learn a different visual representation from the convolutional neural network. They claimed ViTs are less biased to texture, which is assumed to be suspicious signals in the image, and show promising results under image corruption. We find vision transformers can serve as a strong baseline in domain generalization tasks, and we take advantage of the unique properties of ViTs to develop the~\ours~algorithm.

\paragraph{Domain generalization algorithms.} Many efforts~\cite{li_domain_2018,wang_heterogeneous_2020,li_learning_2017,sun_deep_2016,sagawa_distributionally_2020,ganin_domain-adversarial_2016,li_deep_2018,arjovsky_invariant_2020} have been devoted to learning a model that can generalize out-of-domains by the research community.
We follow the categorization in~\cite{zhou_domain_2021} to overview previous DG algorithm and select the representative ones to be benchmarked with the vision transformer in Appendix~\ref{sec:dgmethods}.

\section{Method}

In this section, we first preliminary knowledge about the domain generalization task and the vision transformer. Then, an overview of~\ours~algorithm is provided. After that, two parts of the algorithm, Domain Prompt Learning (DPL) and Prompt Adapter Learning (PAL) are presented.

\subsection{Preliminary}

\paragraph{Problem Setting.} In domain generalization, the training dataset is composed of $K$ source domains where each domain $\mathcal{D}_i=\{({\bm x}_i,y_i)\}_{i=1}^{n_i}$, where ${\bm x}\in\mathcal{X}$ denotes the image and $y\in\mathcal{Y}$ is the label. The whole source dataset can be represented as, $\mathcal{D}_S=\bigcup_{i=1}^K\mathcal{D}_i=\{({\bm x}_i,y_i,d_i)\}_{i=1}^{n_s}$ where $d_i$ stands for the domain label of an image and $n_s = \sum_{i=1}^Kn_i$.  
DG goal is to learn a model $F:\mathcal{X}\rightarrow Y$ that can generalize well to a target domain $\mathcal{D}_T$, which is not accessible during training. Typically, the predictor $F$ can be decomposed into $\phi\circ f$, where the feature extractor $f:\mathcal{X}\rightarrow\mathcal{H}$ learns the representation of the input image, and then the classifier $\phi:\mathcal{H}\rightarrow\mathcal{Y}$ predicts the class. In our case, the feature extractor $f$ is a vision transformer.

\paragraph{Vision Transformer.} The input image to a vision transformer $\bm x$ is first divided into $k$ patches $\{I_i\}_{i=1}^k$. The encoding layer $E$ transforms the input patches into patch tokens, and positional embeddings are added to them. The input to the transformer is the encoded patch tokens plus a classification token $\clstoken$. The vision transformer consists of $\ell$ blocks, and each block contains an attention layer and an MLP layer. The prediction of the vision transformer can be formulated as follows:
\begin{align}
    \clstoken &= f([\clstoken, E(I_1), \cdots, E(I_k)) \\
    y &= \phi(\clstoken),
\end{align}
where $[\cdot]$ means concatenation of tokens. The interaction of different tokens happens in the attention layers of the vision transformer.

\begin{figure}[t]
\begin{center}
\centerline{\includegraphics[width=\textwidth]{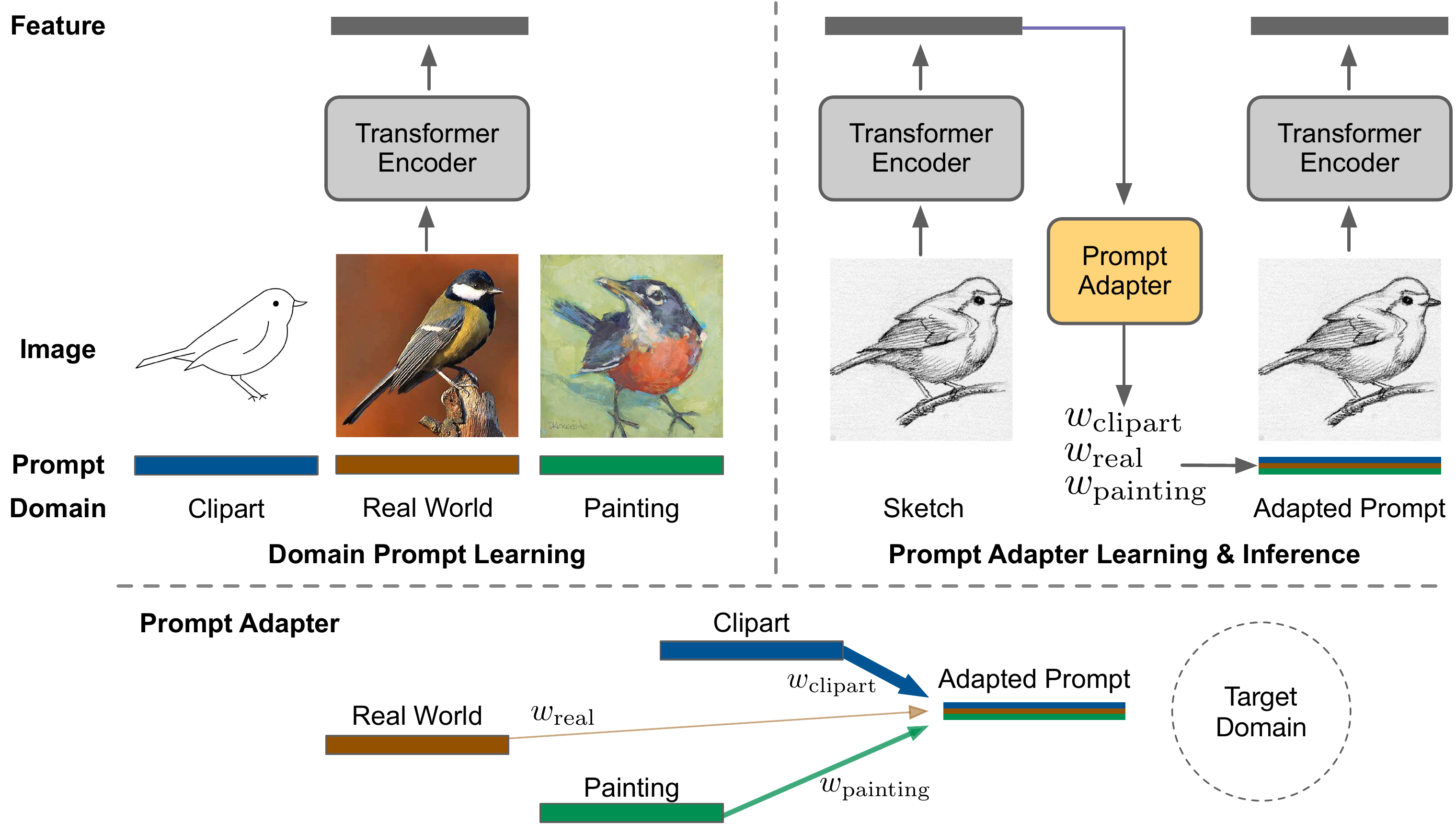}}
\caption{{\bf An overview of the proposed \ours~algorithm}. {\bf Left} shows the Domain Prompt Learning. {\bf Right} shows both the Prompt Adapter Learning and the inference process. {\bf Bottom} shows adapted prompt generation process, where each domain prompt is placed according to the domain position in the feature space. A thicker line denotes a larger weight.}
\label{fig:framework}
\end{center}
\end{figure}

\subsection{Overview}

In domain generalization, our goal is to learn a robust model that can be applied to an unseen domain with multiple source domains. A trivial way is to learn one feature extractor to embed images from all source domains into a common feature space. With a shared featurizer and classifier across all domains, we can train the model with Empirical Risk Minimization (ERM) loss as the following:
\begin{equation}
    \mathcal{L}_{\text{ERM}}=\frac{1}{K}\sum_{i=1}^K\frac{1}{n_i}\sum_{(x,y)\in\mathcal{D}_i}\ce(\phi(f({\bm x})), y).
\end{equation}
However, the network is ignorant of the domain difference if we optimize the above loss directly on all domains. Our motivation is to learn better models for each domain by taking advantage of the domain information and properly exploiting them to perform robust prediction in the target domain. Domain-specific batchnorm~\cite{Seo2020LearningTO} uses different batch normalization statistics and affine parameters for each domain. By doing this, the network with one specific batch normalization is optimized for the corresponding domain.

While the domain-specific structure can enhance the generalization performance of the model, 
there are some limitations of it. First, the domain information is limited to some specific layers. The network can only learn different normalizations for each domain. Second, when predicting in a target domain, the network cannot tell which domain-specific batchnorm is the most suitable one to use. As a result, a simple average of logits predicted by the model with different domain batchnorms is used for prediction. However, the distance between the target domain and source domains are different, as shown in Table~\ref{tab:dist}, and a simple average may lead to bad performance.

In our method, we propose Domain Prompt Learning (DPL) and Prompt Adapter Learning (PAL) to learn a more discriminative feature in each source domain and a more robust combination of knowledge learned from different source domains for inference. Figure~\ref{fig:framework} shows the framework of our method. The left part shows learning with domain prompts (see \ref{subsec:dpl}), where the image is input combined with corresponding domain tokens. The right part shows both the Prompt Adapter Learning in training time and the inference method (see \ref{subsec:pal}). The domain adapter gives a combination of learned source domain prompts by image features extracted with ViT backbone. The output adapted prompt is optimized for each input image.

\subsection{Domain Prompt Learning}
\label{subsec:dpl}

To learn network optimized for each source domain, we introduce the domain prompts to carry the domain-specific knowledge. For each domain, we have a set of specific prompts, namely $L$ prompt tokens. The prompt collections for $K$ source domains are:
\begin{equation}
    \texttt{prompts} = \Big\{\{\texttt{[DOM]}^j\}_{i=1}^L\Big\}_{j=1}^K.
\end{equation}
Following the common practice in nature language processing, we append the prompts with the other tokens. With the domain prompt, the domain-aware feature is extracted as:
\begin{equation}
    \clstoken = f_d({\bm x}) = f([\,\,\clstoken,\,\,\{E(I_i)\}_{i=1}^k,\,\,\texttt{prompts}_d\,\,]),
\end{equation}
where $\texttt{prompts}_d$ means the prompts corresponding to domain $d$, and $f_d$ represents the original model with appended domain prompts. The $\texttt{prompts}$ are learnable parameters to capture knowledge contributing to a correct prediction on each domain. The training loss for DAL is:
\begin{equation}
    \mathcal{L}_{\text{prompt}} =\frac{1}{|\mathcal{D}_S|}\sum_{({\bm x},y,d)\in\mathcal{D}_S}\ce(\phi(f_d({\bm x})), y).\label{eq:dpl}
\end{equation}
In vision transformer, domain prompts are treated the same as image tokens and the class token. They pass all attention layers and MLP layers. The domain knowledge is utilized through the attention mechanism, where image and class tokens can get information from the domain tokens. In this way, the whole model can take advantage of the domain information instead of some specific layers to give a more accurate prediction.

\begin{algorithm}[t]
\caption{\ours~Algorithm}
\begin{algorithmic}[1]
    \REQUIRE number of steps $T$, featurizer $f$, classifier $\phi$, domain prompts $\texttt{prompts}$, domain adapter $A$
    \FOR{$s \gets 1$ to $T$}
        \STATE Sample $(x^{i_1}_1, y^{i_1}_1),\cdots,(x^{i_m}_m, y^{i_m}_m)$ from $m$ domains $\mathcal{D}_1,\cdots,\mathcal{D}_m$
        \STATE $\rhd$ Learning with domain prompts
        \STATE Calculate loss $\mathcal{L}_{\text{prompt}}$ according to Equation~\ref{eq:dpl}
        \STATE $\rhd$ Learning with domain adapter
        \FOR{each domain $i$}
            \STATE $h_i \gets \text{stop\_gradients}(f(x_i))$
            \STATE $\texttt{adapted\_prompts}\gets A(h)$
        \ENDFOR
        \STATE Calculate loss $\mathcal{L}_{\text{w}}, \mathcal{L}_{\text{adapt}} $ according to Equation~\ref{eq:pal} and Equation~\ref{eq:la}
        \STATE $\mathcal{L} \gets \mathcal{L}_{\text{prompt}}+\mathcal{L}_{\text{adapt}}+\lambda\mathcal{L}_{\text{w}}$
        \STATE Backward $\mathcal{L}$ and update the parameters
    \ENDFOR
\end{algorithmic}
\label{alg}
\end{algorithm}

\subsection{Inference and Prompt Adapter Learning}
\label{subsec:pal}

With Domain Prompt Learning, we can learn a good feature extractor for our source domains. As domain generalization asks for prediction on unseen target domains, it is important to design an approach to extend our knowledge from source domains. One direct method is to average the logits derived by different domain prompts, \ie~$\sum_{d=1}^{n_s}\phi(f_d(x))$. However, this method is ignorant of the relationship between different domains. As shown in Table~\ref{tab:dist}, the distances between different domains vary, so we hope to exploit the knowledge in domain prompts while considering this information.

Since each domain prompt is optimized for one specific domain, we believe that for an input image, there also exists an optimal prompt for the domain the image comes from. As we only have domain knowledge of the source domains, we represent the prompt for unknown domains as a linear combination of different source domains:
\begin{equation}
    \texttt{adapted\_prompt} = A(f({\bm x})) = \Big\{\sum_{d=1}^{K}w_d^j\cdot\texttt{[DOM]}_d^j\Big\}_{j=1}^K.
\end{equation}


We propose a prompt adapter $A$ to predict the linear combination weight $w_d$ based on the extracted feature. Without domain prompts, the extracted features are more domain-discriminative (see Table~\ref{tab:dodist} row 3), and $A$ is able to learn the relationship between different domains and input images. We set $A$ to be two-layer MLP with a appended softmax layer to ensure $\sum_{d=1}^{K}w_d^j=1$. The combination is visualized in Figure~\ref{fig:framework} bottom. The prompt adapter learns an adaptive weight by analyzing the relationship between input target image and the source domains. It can automatically assign a larger weight for the domain prompt whose domain is closer to the target one than the others.

The inference process for any input image from unseen domains can be formulated as follows:
\begin{align}
    \texttt{[CLS]}' &= f([\,\,\clstoken,\,\,\{E(I_i)\}_{i=1}^k]) \label{eq:inf_st}\\
    \texttt{adapted\_prompts} &= A(\,\,\texttt{[CLS]}'\,\,) \\
    \texttt{[CLS]}'' &= f([\,\,\texttt{[CLS]}',\,\,\{E(I_i)\}_{i=1}^k,\,\,\texttt{adapted\_prompts}\,\,]) \\
    y &= \phi(\,\,\,\texttt{[CLS]}''\,). \label{eq:inf_end}
\end{align}
The adapted prompts are conditioned on the input image so that a more appropriate prompt can be used for the prediction.


To train the domain adapter, we model the inference process during training by treating each training image an out-of-distribution image. Two losses are imposed in this process to learn a good prompt adapter. First, given the image from domain $t$, the most suitable prompts should be its corresponding domain prompts. Thus, we use a cross-entropy loss directly on the predicted weight by $A(f(x))$ as follows:
\begin{equation}
    \mathcal{L}_{\text{w}} = \frac{1}{K}\sum_{j=1}^K\frac{1}{K}\Big(\ce(w_j^j, 1) + \sum_{d\neq t}\ce(w_d^j, 0)\Big). \label{eq:pal}
\end{equation}
On the other hand, the prompt adapter should learn to generate a prompt that can contribute to the final prediction of the image. No matter which domain the image comes from, prompts of other domains may also help in its prediction. Thus, we optimize the ERM loss when the image is input with the adapted prompts:
\begin{equation}
    \mathcal{L}_{\text{adapt}} = \frac{1}{|\mathcal{D}_S|}\sum_{({\bm x},y,d)\in\mathcal{D}_S}\ce(\phi(f_{\texttt{adapted}}({\bm x})), y), \label{eq:la}
\end{equation}
where $f_{\texttt{adapted}}$ is the original model with appended adapted domain prompts.


Taken together, the training algorithm of \ours~is presented in Algorithm~\ref{alg}. Since $\mathcal{L}_{\text{prompt}}$ and $\mathcal{L}_{\text{adapt}}$ are ERM losses, we directly combine them together and use hyper-paramter $\lambda$ to control loss $\mathcal{L}_{\text{w}}$. The inference process follows equations~\ref{eq:inf_st} --~\ref{eq:inf_end}.

\section{Experiment}
\label{sec:experiment}

\subsection{Experimental setting}

\paragraph{Datasets.} We evaluate our approach on four public datasets: PACS, VLCS, OfficeHome, and DomainNet. \textbf{PACS}~\cite{Li2017DeeperBA} is composed of four domains: \underline{\bf P}hotos, \underline{\bf A}rt, \underline{\bf C}artoon, and \underline{\bf S}ketch. It contains 9,991 images in 7 classes. \textbf{VLCS}~\cite{Fang2013UnbiasedML} consists of 10,729 images in 5 classes from 4 real-world datasets, which are \ulbf{V}OC2007, \underline{\bf L}abelMe, \underline{\bf C}altech, and \underline{\bf S}UN09. \textbf{OfficeHome}~\cite{Venkateswara2017DeepHN} is a more difficult dataset as it has 30,475 images in 65 classes. There are four different domains in this dataset: \underline{\bf A}rt, \underline{\bf C}lipart, \underline{\bf P}roduct, and \underline{\bf R}eal. \textbf{DomainNet}~\cite{Peng2019MomentMF} is a large-scale benchmark with 586k images in 345 classes. It contains six domains which are \underline{\bf C}lipart, \underline{\bf I}nfograph, \underline{\bf P}ainting, \underline{\bf Q}uickdraw, \underline{\bf R}eal, and \underline{\bf S}ketch. Each domain is selected once as the target domain for each dataset, while the rest as source domains. Thus, the number of experiments carried out is the same as the number of all domains.

\paragraph{Implementation details.} To fairly compare different methods, following~\cite{gulrajani_search_2020,arpit_ensemble_2021}, we control the same computing condition and budgets. However, \cite{gulrajani_search_2020} conduct a random search of 20 trials on each domain of datasets. With three repeated trials, this results in a total of 1080 experiments to evaluate one method, which is too expensive. Another drawback of the evaluation is that it is hard to choose meaningful hyper-parameters for each algorithm. We adopt a two-stage grid-search to determine hyper-parameters for each method for our implementation. First, we search the general hyper-parameters such as learning rate and weight decay with a grid search for each dataset. Then, we fix these parameters and conduct a grid search for different hyper-parameters for each method. Following~\cite{gulrajani_search_2020}, we train 5k iterations for each setting.

The hyper-parameter search spaces of stage one and stage two are provided in the Appendix~\ref{sec:hp}. Since the hyper-parameter required for the vision transformer may significantly differ from the one for ResNet, we sweep a wide scale of learning rate and weight decay. 
After the grid search, we use a dropout of 0.1 and weight decay of $1e^{-2}$. The learning rate is $5e^{-6}$ for PACS and VLCS, $1e^{-5}$ for OfficeHome, and $5e^{-5}$ for DomainNet.
For~\ours, there are two hyper-parameters: the length of prompts $L$ is set to $4$ according to experimental performance, and the $\lambda$ is searched within $\{0.1, 1, 10\}$.

We conduct our experiments based on the DomainBed~\cite{gulrajani_search_2020} implementation. The framework is based on PyTorch~\cite{Paszke2019PyTorchAI}, and we use ViT-Base/16 model pretrained on ImageNet for all experiments. We train the model with AdamW optimizer~\cite{Kingma2015AdamAM,Loshchilov2019DecoupledWD} and the same data augmentation policy as~\cite{gulrajani_search_2020}. 
We adopt the training-domain validation set method~\cite{gulrajani_search_2020} for model selection. Specifically, each source domain is split into training and validation subsets. For \ours, we apply the same inference process for validation like the one for the test instead of using the corresponding domain tokens. In this way, the validation result is a more reliable indicator of the test performance.

\begin{table}[t]
\centering
\caption{Accuracy (\%) on PACS, VLCS, OfficeHome and DomainNet with ViT-Base/16. The last column is the relative improvement of each method compared to ERM ViT baseline.}
\begin{tabular}{lcccccc}
\toprule
\textbf{Algorithm} & \textbf{PACS} & \textbf{VLCS} & \textbf{OfficeHome} & \textbf{DomainNet} & \textbf{Avg}. & $\Delta$ \textbf{Avg}. \\
\midrule
ERM~\cite{gulrajani_search_2020} (ResNet-50) & 85.7 & 77.4 & 67.5 & 41.2 & 68.0 &  \XY{-3.8}\\
Best in~\cite{gulrajani_search_2020} (ResNet-50) & 86.0 & 77.7 & 68.6 & 41.8 & 68.5 &  \XY{-3.3}\\
\hline\Tstrut
ERM~\cite{gulrajani_search_2020}     & 86.4 & 79.2 & 74.3 & 47.4 & 71.8 & 0.0 \\
IRM~\cite{arjovsky_invariant_2020}    & 83.9 & 79.5 & 72.3 & 27.1 & 65.7 & \XY{-6.1} \\
CDANN~\cite{li_deep_2018} & 82.3 & 79.1 & 72.3 & 34.3 & 67.0 & \XY{-4.8} \\
DANN~\cite{ganin_domain-adversarial_2016} & 82.6 & \underline{79.6} & 71.9 & 35.0 & 67.3 & \XY{-4.5} \\
GDRO~\cite{sagawa_distributionally_2020} & 86.6 & 78.9 & 74.5 & 41.3 & 70.3 & \XY{-1.5} \\
CORAL~\cite{sun_deep_2016} & 86.2 & 79.2 & 74.5 & \underline{47.7} & 71.9 & \FG{+0.1}   \\
MLDG~\cite{li_learning_2017} & 87.0 & 78.7 & \underline{74.8} & 47.6 & 72.0 & \FG{+0.2} \\
Mixup~\cite{wang_heterogeneous_2020} & \underline{87.4} & 79.1 & 74.5 & 46.7 & 72.0 & \FG{+0.2} \\
MMD~\cite{li_domain_2018} & 86.9 & 79.8 & 74.5 & 47.4 & \underline{72.2} & \FG{+0.4}    \\
\hline
\rowcolor{Gray}
\ours & {\bf 88.1} & {\bf 80.4} & {\bf 76.0} & {\bf 48.3} & {\bf 73.2} & \FG{\bf +1.4} \\
\bottomrule
\end{tabular}
\label{tab:full}
\end{table}

\begin{table}[!t]
\centering
\caption{Averaged domain distances with different models.}
{\small 
\begin{tabular}{lccc}
\toprule
Model & cross/in dist. & cross/in avg. class dist. \\
\midrule
ResNet-50 (ImageNet Pretrained) & 43 & 114  \\
ViT-B/16 (ImageNet Pretrained) & 14 & 17 \\ \hline\Tstrut
\ours~w/o. domain prompts & 12 & 10 \\
\ours~w/o. fine-tuning backbone & 21 & 41 \\
\ours & 6 & 2 \\
\bottomrule
\end{tabular}
}
\label{tab:dodist}
\end{table}

\begin{table}[!t]
\centering
\caption{Distance between different domains in the OfficeHome dataset on features from pretrained ResNet-50 (left) and ViT-Base/16 (right). In each table, upper right values are the distance between domains while lower left values denote the averaged distance between classes in different domains.}
\begin{subtable}{.47\linewidth}
    \centering
\begin{tabular}{ccccc}
\toprule
 & \ulbf{A}rt & \ulbf{C}lipart & \ulbf{P}ainting & \ulbf{R}eal \\
\midrule
\textbf{A} & -- & 69.7 & 37.1 & 14.0 \\
\textbf{P} & \textit{73.3} & -- & 54.1 & 69.7 \\
\textbf{R} & \textit{178.8} & \textit{159.6} & -- & 17.6 \\
\textbf{C} & \textit{124.2} & \textit{99.7} & \textit{51.6} & --\\
\bottomrule
\end{tabular}
\end{subtable}
\begin{subtable}{.49\linewidth}
    \centering
\begin{tabular}{ccccc}
\toprule
 & \ulbf{A}rt & \ulbf{C}lipart & \ulbf{P}ainting & \ulbf{R}eal \\
\midrule
\textbf{A} & -- & 12.4 & 11.5 & 7.0 \\
\textbf{P} & \textit{10.6} & -- & 23.5 & 22.1 \\
\textbf{R} & \textit{38.4} & \textit{26.3} & -- & 7.1 \\
\textbf{C} & \textit{13.8} & \textit{10.5} & \textit{5.8} & --\\
\bottomrule
\end{tabular}
\end{subtable}
\label{tab:dist}
\end{table}

\paragraph{Domain distance.} The difficulty of domain generalization lies in the domain shift across different domains. A well-learned feature space should have images with the same label close and images with different classes distant. However, due to large distribution discrepancy across
domains, features of images in the same domain tend to have a smaller distance. We describe this property by cosine distance of domain centroids (detailed in Appendix~\ref{sec:dcd}). The distance is measured on features from a ImageNet pretrained network.
The domain distance measures the distance between two domain centroids, and the average domain class distance is the average class distance between different domains. 
Table~\ref{tab:dist} shows the two distances between different domains, and Table~\ref{tab:dodist} shows the two distances averaged on different domains.

\subsection{Domain generalization results}

First, we benchmark previous methods with ViT-base/16 backbone on PACS, VLCS, OfficeHome, and DomainNet, with the averaged accuracy of four datasets in Table \ref{tab:full}, and the results for each dataset in Table~\ref{tab:offdn} and Table~\ref{tab:pacsvlcs}. 
Unless otherwise specified, the bold text denotes the best results, and the underlined one is the best except for our methods. The ERM method on ViT-base/16 outperforms the ERM method, and the best results trained with ResNet-50 as the backbone by 3.8\% and 3.3\%, respectively. One reason for this phenomenon can be found in Table~\ref{tab:dist} and Table~\ref{tab:dodist}. The domain distance of features from ViT is much smaller than the one from ResNet-50 pretrained on ImageNet. Another observation is that as DG methods only improve the ERM baseline by 0.5\% with ResNet-50, they also achieve a slight improvement with the ViT-base backbone. Only 0.4\% averaged accuracy is achieved by the best previous methods on the four public datasets.

We compare our proposed \ours~with previous methods. \ours~improves the ERM baseline with ViT backbone by 1.4\% while all other methods have an improvement of no more than 0.4\% against the ERM baseline, which is a 3.5$\times$ improvement. \ours~achieves the best accuracy on each of the four benchmark datasets. If we look at the result of each target domain setting in each dataset, our algorithm outperforms previous methods on 16 out of 18 settings.

\begin{table}[!t]
    \caption{Accuracy (\%) on OfficeHome (left) and DomainNet (right) with ViT-Base/16. }
    \begin{subtable}{.425\linewidth}
        \centering
        \resizebox{\columnwidth}{!}{
            \begin{tabular}{lccccc}
            \toprule
             & \textbf{A} & \textbf{C} & \textbf{P} & \textbf{R} & \textbf{Avg}. \\ 
            \midrule
            ERM~\cite{gulrajani_search_2020} & 71.4$_{\pm 0.0}$ & 59.4$_{\pm 0.5}$ & 81.6$_{\pm 0.1}$ & \underline{84.9$_{\pm 0.3}$} & 74.3 \\
            MMD~\cite{li_domain_2018} & 71.8$_{\pm 0.2}$ & 60.7$_{\pm 1.0}$ & \underline{81.4$_{\pm 0.3}$} & 84.3$_{\pm 0.1}$ & \underline{74.5} \\
            CORAL~\cite{sun_deep_2016} & 72.3$_{\pm 0.7}$ & \underline{60.9}$_{\pm 1.0}$ & 80.4$_{\pm 0.4}$ & 84.3$_{\pm 0.0}$ & \underline{74.5} \\
            DANN~\cite{ganin_domain-adversarial_2016} & 68.7$_{\pm 0.9}$ & 55.8$_{\pm 0.4}$ & 79.2$_{\pm 0.0}$ & 83.9$_{\pm 0.2}$ & 71.9 \\
            CDANN~\cite{li_deep_2018} & 68.2$_{\pm 0.4}$ & 56.8$_{\pm 0.4}$ & 79.8$_{\pm 0.4}$ & 84.2$_{\pm 0.0}$ & 72.3 \\
            MLDG~\cite{li_learning_2017} & 72.3$_{\pm 0.0}$ & 60.0$_{\pm 1.1}$ & 82.2$_{\pm 0.2}$ & 84.7$_{\pm 0.1}$ & 74.8 \\
            GDRO~\cite{sagawa_distributionally_2020} & 71.2$_{\pm 0.7}$ & \underline{60.9}$_{\pm 0.4}$ & 81.1$_{\pm 0.3}$ & 84.7$_{\pm 0.3}$ & \underline{74.5}\\
            IRM~\cite{arjovsky_invariant_2020} & 69.5$_{\pm 0.1}$ & 57.1$_{\pm 0.3}$ & 79.1$_{\pm 0.1}$ & 83.3$_{\pm 0.1}$ & 72.3 \\
            Mixup~\cite{wang_heterogeneous_2020} & \underline{72.7}$_{\pm 0.4}$ & 59.7$_{\pm 0.5}$ & 80.9$_{\pm 0.3}$ & 84.6$_{\pm 0.2}$ & \underline{74.5}\\
            \rowcolor{Gray}
            \ours & {\bf 72.7}$_{\pm 0.5}$ & {\bf 62.3}$_{\pm 0.6}$ & {\bf 83.2}$_{\pm 0.1}$ & {\bf 85.9}$_{\pm 0.1}$ & {\bf 76.0} \\
            \bottomrule
            \end{tabular}
        }
    \end{subtable}%
    \begin{subtable}{.575\linewidth}
        \centering
        \resizebox{\columnwidth}{!}{
            \begin{tabular}{lccccccc}
            \toprule
             & \textbf{clip} & \textbf{info} & \textbf{paint} & \textbf{quick} & \textbf{real} & \textbf{sketch} & \textbf{Avg}. \\
            \midrule
            ERM~\cite{gulrajani_search_2020} & \underline{67.0$_{\pm 0.1}$} & 23.4$_{\pm 0.6}$ & 54.5$_{\pm 0.3}$ & 15.8$_{\pm 0.6}$ & \underline{69.3}$_{\pm 0.4}$ & 54.6$_{\pm 0.5}$ & 47.4 \\
            MMD~\cite{li_domain_2018} & \underline{67.0}$_{\pm 0.2}$ & \underline{23.8}$_{\pm 0.2}$ & 54.0$_{\pm 0.0}$ & 15.9$_{\pm 0.5}$ & 69.0$_{\pm 0.1}$ & 54.6$_{\pm 0.2}$ & 47.4 \\
            CORAL~\cite{sun_deep_2016} & 66.8$_{\pm 0.2}$ & 24.4$_{\pm 0.1}$ & 54.6$_{\pm 0.2}$ & 16.2$_{\pm 0.2}$ & \underline{69.3}$_{\pm 0.1}$ & \underline{54.9} $_{\pm 0.3}$ & \underline{47.7} \\
            DANN~\cite{ganin_domain-adversarial_2016} & 45.6$_{\pm 0.1}$ & 14.4$_{\pm 0.2}$ & 44.6$_{\pm 0.6}$ & 8.1$_{\pm 0.0}$ & 52.3$_{\pm 0.3}$ & 44.8$_{\pm 0.5}$ & 35.0 \\
            CDANN~\cite{li_deep_2018} & 45.3$_{\pm 0.5}$ & 13.5$_{\pm 0.0}$ & 45.8$_{\pm 0.0}$ & 7.7$_{\pm 0.2}$ & 49.6$_{\pm 0.1}$ & 43.8$_{\pm 0.2}$ & 34.3 \\
            MLDG~\cite{li_learning_2017} & 65.8$_{\pm 1.4}$ & \underline{23.8}$_{\pm 0.1}$ & \underline{54.7}$_{\pm 0.1}$ & \underline{17.2}$_{\pm 0.2}$ & 69.0$_{\pm 0.1}$ & \underline{54.9}$_{\pm 0.3}$ & 47.6 \\
            GDRO~\cite{sagawa_distributionally_2020} & 59.3$_{\pm 0.2}$ & 20.8$_{\pm 0.2}$ & 46.0$_{\pm 0.6}$ & 11.3$_{\pm 0.5}$ & 63.7$_{\pm 0.1}$ & 47.0$_{\pm 0.3}$ & 41.3\\
            IRM~\cite{arjovsky_invariant_2020} & 35.6$_{\pm 0.5}$ & 13.9$_{\pm 0.2}$ & 33.1$_{\pm 0.8}$ & 6.4$_{\pm 0.2}$ & 40.6$_{\pm 0.4}$ & 32.8$_{\pm 0.3}$ & 27.1 \\
            Mixup~\cite{wang_heterogeneous_2020} & 65.1$_{\pm 0.1}$ & 23.7$_{\pm 0.1}$ & 54.4$_{\pm 0.1}$ & 15.1$_{\pm 0.3}$ & 67.9$_{\pm 0.0}$ & 54.0$_{\pm 0.1}$ & 46.7\\
            \rowcolor{Gray}
            \ours & {\bf 67.7}$_{\pm 0.2}$ & {\bf 24.6}$_{\pm 0.1}$ & {\bf 54.9}$_{\pm 0.1}$ & {\bf 17.5}$_{\pm 0.2}$ & {\bf 69.6}$_{\pm 0.3}$ & {\bf 55.2}$_{\pm 0.5}$ & {\bf 48.3} \\
            \bottomrule
            \end{tabular}
        }
    \end{subtable}%
    \label{tab:offdn}
\end{table}

\begin{table}[!t]
    \caption{Accuracy (\%) on PACS (left) and VLCS (right) with ViT-Base/16. }
    \begin{subtable}{.5\linewidth}
        \centering
        \resizebox{\columnwidth}{!}{
            \begin{tabular}{lccccc}
            \toprule
             & \textbf{A} & \textbf{C} & \textbf{P} & \textbf{S} & \textbf{Avg}. \\ 
            \midrule
            ERM~\cite{gulrajani_search_2020} & 88.6$_{\pm 0.3}$ & 81.3$_{\pm 0.3}$ & 99.1$_{\pm 0.1}$ & 76.1$_{\pm 0.9}$ & 86.4 \\
            MMD~\cite{li_domain_2018} & 89.9$_{\pm 0.5}$ & 80.8$_{\pm 0.5}$ & 98.7$_{\pm 0.1}$ & \underline{77.1}$_{\pm 0.7}$ & 86.6 \\
            CORAL~\cite{sun_deep_2016} & 89.7$_{\pm 0.3}$ & 80.2$_{\pm 1.0}$ & 98.7$_{\pm 0.2}$ & 76.1$_{\pm 1.6}$ & 86.2 \\
            DANN~\cite{ganin_domain-adversarial_2016} & 87.5$_{\pm 0.4}$ & 78.6$_{\pm 0.9}$ & 98.0$_{\pm 0.2}$ & 66.3$_{\pm 1.7}$ & 82.6 \\
            CDANN~\cite{li_deep_2018} & 87.3$_{\pm 0.6}$ & 80.1$_{\pm 0.5}$ & 98.5$_{\pm 0.2}$ & 63.1$_{\pm 2.9}$ & 82.3 \\
            MLDG~\cite{li_learning_2017} & 90.4$_{\pm 0.5}$ & 83.0$_{\pm 0.7}$ & 98.7$_{\pm 0.0}$ & 76.0$_{\pm 0.2}$ & 87.0 \\
            GDRO~\cite{sagawa_distributionally_2020} & 90.1$_{\pm 0.6}$ & 80.5$_{\pm 1.0}$ & 98.8$_{\pm 0.1}$ & 77.1$_{\pm 1.5}$ & 86.6\\
            IRM~\cite{arjovsky_invariant_2020} & 89.1$_{\pm 0.7}$ & 79.1$_{\pm 1.1}$ & 98.3$_{\pm 0.1}$ & 68.2$_{\pm 5.1}$ & 83.9 \\
            Mixup~\cite{wang_heterogeneous_2020} & \underline{91.1}$_{\pm 0.1}$ & \underline{84.0}$_{\pm 0.3}$ & \underline{99.2}$_{\pm 0.2}$ & 75.4$_{\pm 0.5}$ & \underline{87.4}\\
            \rowcolor{Gray}
            \ours & {\bf 91.1}$_{\pm 0.3}$ & 83.0$_{\pm 0.2}$ & {\bf 99.6}$_{\pm 0.0}$ & {\bf 78.7}$_{\pm 0.8}$ & {\bf 88.1} \\
            \bottomrule
            \end{tabular}
        }
        \label{tab:pacs}
    \end{subtable}%
    \begin{subtable}{.5\linewidth}
        \centering
        \resizebox{\columnwidth}{!}{
            \begin{tabular}{lccccc}
            \toprule
             & \textbf{C} & \textbf{L} & \textbf{S} & \textbf{V} & \textbf{Avg}. \\
            \midrule
            ERM~\cite{gulrajani_search_2020} & 96.9$_{\pm 0.3}$ & 65.3$_{\pm 0.4}$ & 75.2$_{\pm 0.5}$ & 79.6$_{\pm 1.4}$ & 79.2 \\
            MMD~\cite{li_domain_2018} & 96.8$_{\pm 0.1}$ & 65.6$_{\pm 1.0}$ & \underline{77.3}$_{\pm 0.6}$ & \underline{79.3}$_{\pm 1.5}$ & \underline{79.8} \\
            CORAL~\cite{sun_deep_2016} & 96.2$_{\pm 0.7}$ & 66.0$_{\pm 0.3}$ & 76.8$_{\pm 0.6}$ & 78.0$_{\pm 0.9}$ & 79.2 \\
            DANN~\cite{ganin_domain-adversarial_2016} & \underline{97.7}$_{\pm 0.3}$ & \underline{66.5}$_{\pm 0.5}$ & 76.0$_{\pm 1.0}$ & 78.0$_{\pm 0.7}$ & 79.6 \\
            CDANN~\cite{li_deep_2018} & 96.3$_{\pm 0.2}$ & 66.3$_{\pm 0.5}$ & 76.5$_{\pm 0.6}$ & 77.3$_{\pm 0.5}$ & 79.1 \\
            MLDG~\cite{li_learning_2017} & 95.7$_{\pm 0.5}$ & 65.9$_{\pm 0.3}$ & 76.3$_{\pm 0.8}$ & 77.1$_{\pm 0.2}$ & 78.7 \\
            GDRO~\cite{sagawa_distributionally_2020} & 97.1$_{\pm 0.5}$ & 65.2$_{\pm 0.7}$ & 75.1$_{\pm 1.0}$ & 78.3$_{\pm 1.0}$ & 78.9\\
            IRM~\cite{arjovsky_invariant_2020} & 96.2$_{\pm 0.5}$ & 65.7$_{\pm 0.8}$ & 76.8$_{\pm 0.4}$ & 77.7$_{\pm 0.9}$ & 79.5 \\
            Mixup~\cite{wang_heterogeneous_2020} & 97.1$_{\pm 0.9}$ & 65.4$_{\pm 0.2}$ & 76.1$_{\pm 0.4}$ & 78.0$_{\pm 0.3}$ & 79.1\\
            \rowcolor{Gray}
            \ours & 97.2$_{\pm 0.1}$ & {\bf 67.4}$_{\pm 0.5}$ & {\bf 77.4}$_{\pm 1.0}$ & {\bf 79.7}$_{\pm 0.5}$ & {\bf 80.4}\\
            \bottomrule
            \end{tabular}
        }
    \end{subtable}%
    \label{tab:pacsvlcs}
\end{table}

\begin{table}[!t]
\centering
\caption{Ablation study of \ours~on OfficeHome.}
{\small
\begin{tabular}{lccccc}
\toprule
\textbf{Algorithm} & \textbf{A} & \textbf{C} & \textbf{P} & \textbf{R} & \textbf{Avg}. \\
\midrule
\ours & {\bf 72.7}$_{\pm 0.5}$ & {\bf 62.3}$_{\pm 0.6}$ & {\bf 83.2}$_{\pm 0.1}$ & {\bf 85.9}$_{\pm 0.1}$ & {\bf 76.0} \\
w/o. Domain Prompts & 71.4$_{\pm 0.0}$ & 59.4$_{\pm 0.5}$ & 81.6$_{\pm 0.1}$ & 84.9$_{\pm 0.3}$ & 74.3 \\ 
w/o. Domain Adapter & 71.7$_{\pm 0.3}$ & 60.9$_{\pm 0.5}$ & 81.9$_{\pm 0.3}$ & 84.6$_{\pm 0.2}$ & 74.8 \\ \hline
\Tstrut
w/o. $\mathcal{L}_{\text{w}}$ & 72.1$_{\pm 0.1}$ & 62.1$_{\pm 0.3}$ & 82.8$_{\pm 0.5}$ & 84.9$_{\pm 0.3}$ & 75.5 \\
w/o. $\mathcal{L}_{\text{adapt}}$ & 71.8$_{\pm 0.1}$ & 60.9$_{\pm 0.5}$ & 82.1$_{\pm 0.2}$ & 84.1$_{\pm 0.1}$ & 74.7\\ \hline\Tstrut
w/o. Fine-tuning backbone & 71.2$_{\pm 0.3}$ & 58.7$_{\pm 0.5}$ & 81.9$_{\pm 0.2}$ & 84.1$_{\pm 0.2}$ & 74.0 \\
\bottomrule
\end{tabular}
}
\label{tab:ablation}
\end{table}

\subsection{Ablation study}

We investigate the effectiveness of different components in \ours~on OfficeHome. Table~\ref{tab:ablation} shows the performance of different variants of our design. For the version without a domain adapter, since we cannot generate the adapted prompt, we use the averaged logits from different domain prompts for inference. Our two learning objectives of \ours, Domain Prompt Learning and Prompt Adapter Learning, are both vital to the final performance. The great performance drop shows that the two components are both effective. For training the domain adapter, we show that our two losses are both effective. The results in rows 4 and 5 show that each loss is contributive to our final algorithm.

In our experiments, we fine-tune the whole model instead of only the prompt as in the natural language processing. Training only the prompts results in much lower accuracy in Table~\ref{tab:ablation}. One reason for this is that the number of parameters of the prompts is not enough to fully learn the downstream tasks. Another reason we find is shown in Table~\ref{tab:dodist}. Tuning prompts only cannot 
reduce the distance between different domains, thus resulting in a bad DG performance.

\subsection{Domain adapter learns good prompts}

To verify that our prompt adapter can learn a good prompt for the unseen domain prediction, we test the DG performance in the target domain with different domain prompts. Table~\ref{tab:diff} shows the target accuracy when predicted with the adapted prompt or a domain prompt from one source domain. As we can see, the prompt generated by the prompt adapter is better than the source domain prompts. This shows the prompt adapter does learn how to generate a good combination of domain prompts for generalized prediction.

Table~\ref{tab:adapter} shows the statistics of the weight of adapted prompts generated by the prompt adapter in source and target domains. In the first two tables, each row denotes the classification of one domain, and each column denotes the weight corresponding to each domain prompt. The first table shows the percentage of weights that are the largest in each linear combination weight, and the second one is the averaged weight value for each domain prompt weight. For source domains, the corresponding domain prompts produced by the prompt adapter contain the most information from the corresponding domain as its weight of it is the largest one. This means the prompt adapter learns how to distinguish between different source domains.

The third table shows the domain distance between different domains. For the target domain Clipart, it is most close to the Art domain and far away from the Painting domain. Our prompt adapter learns to give a higher proportion of weights to the Art domain and less to the Painting domain. This reveals the prompt adapter can use the similarity between different domains to produce a better prompt for the target domain during test time.

\begin{table}[t]
\centering
\caption{\ours~test accuracy when predicted with adapted prompt and domain prompts from source domains on OfficeHome dataset.}
{\small
\begin{tabular}{cccccc}
\toprule
\textbf{Target} & \textbf{Adapted} & \textbf{A} & \textbf{C} & \textbf{P} & \textbf{R} \\
\midrule
\textbf{A} & {\bf 72.9} & -- & 72.5 & 72.2 & 72.0 \\
\textbf{C} & {\bf 62.1} & 62.0 & -- & 61.5 & 61.9 \\
\textbf{P} & {\bf 83.0} & 82.8 & 81.7 & -- & 83.0 \\
\textbf{R} & {\bf 85.2} & 84.8 & 84.1 & 84.0 & -- \\
\bottomrule
\end{tabular}
}
\label{tab:diff}
\end{table}

\begin{table}[!t]
\centering
\caption{Prompt adapter weight analysis on OfficeHome dataset with Clipart as the target domain.}
{\small
\begin{tabular}{cccccccccc}
\toprule
 & \multicolumn{3}{c}{\bf Percentage (\%)} & \multicolumn{3}{c}{\bf Average Value} & \multicolumn{3}{c}{\bf Domain Distance}\\\cmidrule(lr){2-4}\cmidrule(lr){5-7}\cmidrule(lr){8-10}
 & \textbf{A} & \textbf{P} & \textbf{R} & \textbf{A} & \textbf{P} & \textbf{R} & \textbf{A} & \textbf{P} & \textbf{R} \\
\midrule
\textbf{A} & 67.3 & 3.5 & 29.2 & 0.62 & 0.08 & 0.30 & -- & 1.5 & 1.1 \\
\textbf{P} & 24.1 & 67.7 & 8.2 & 0.33 & 0.56 & 0.11 & 1.5 & -- & 0.7 \\
\textbf{R} & 3.8 & 7.9 & 88.3 & 0.10 & 0.11 & 0.79 & 1.1 & 0.7 & -- \\
\midrule
\textbf{C} & 70.8 & 3.7 & 25.5 & 0.73 & 0.09 & 0.18 & 4.9 & 9.5 & 7.8 \\
\bottomrule
\end{tabular}
}
\label{tab:adapter}
\end{table}

\section{Conclusion}
In this paper, we first discovered that previous methods only have slight improvement when applied with vision transformers. We propose a novel domain generalization method~\ours~inspired by prompt learning. \ours~learns domain-specific knowledge in each domain and a prompt adapter to generate suitably adapted prompt for prediction on out-of-domain images.
Extensive experiments on multiple datasets demonstrate the superiority of \ours~over previous methods.

\newpage
\printbibliography

\newpage
\appendix
\noindent\textbf{\Large Appendix}

\section{Additional Domain Generalization Methods}
\label{sec:dgmethods}
We follow the categorization in~\cite{zhou_domain_2021} to overview previous DG algorithm and select the representative ones to be benchmarked with the vision transformer.
\begin{itemize}
    \item \textbf{Domain Alignment} methods are based on the assumption that models robust to out-of-domain distribution should learn features invariant to different source domains. These algorithms reduce the distance of learned representations between different domains. {CORAL}~\cite{sun_deep_2016} minimizes the statistic difference while {MMD}~\cite{li_domain_2018} minimizes the Maximum Mean Discrepancy between different domains. {DANN}~\cite{ganin_domain-adversarial_2016} and {CDANN}~\cite{li_deep_2018} are adversarial-based algorithms that use domain discriminator to measure the domain distance, with the latter conditioning on the class label.
    \item \textbf{Meta-Learning} can also help domain generalization. {MLDG}~\cite{li_learning_2017} meta-learns the entire neural network by dividing source domains into nonoverlapping meta-source and meta-target, which simulates the goal of domain generalization to the unseen target domain.
    \item \textbf{Data Augmentation} enlarges the training datasets to cover more unseen distributions. Easy augmentation such as {Mixup}~\cite{wang_heterogeneous_2020} can help boost generalization ability. The most successful augmentation for domain generalization is related to style transfer \cite{Borlino2021RethinkingDG}, where images are transferred into different styles by a style transfer model such as AdaIN~\cite{Huang2017ArbitraryST}. Mixing different styles of images and features is also an effective way~\cite{Nam2021ReducingDG,zhou_domain_2021}. However, these works are based on the assumption that the statistics of the CNN activation map contain the style information, which cannot be directly extended to the vision transformer setting.
    \item \textbf{Ensemble Learning} methods can be generally divided into two groups, one ensembles models trained on different domains~\cite{Seo2020LearningTO,Ding2018DeepDG} while another ensembles models at different steps during training~\cite{Cha2021SWADDG,arpit_ensemble_2021}. The prediction of the ensembled model is believed to be more robust than any single model.
    \item \textbf{Regularization Strategies} are also helpful for predicting out-of-distribution data. Among them, {IRM}~\cite{arjovsky_invariant_2020} forces the linear classifier on each domain should be optimal, and {GroupDRO}~\cite{sagawa_distributionally_2020} increases the weight of harder domains corresponding to higher classification errors.
\end{itemize}


\section{Domain Cosine Distance}
\label{sec:dcd}
To depict the distance between different domains, we measure the distance between different domains by the distance of corresponding centroids. The domain centroid and the class centroid in a domain is defined as follows:
\begin{equation}
    \texttt{cent}_d=\frac{1}{|\mathcal{D}_d|}\sum_{i=1}^{\mathcal{D}_d}{\bm x}_i;\quad \texttt{cent}_c^d=\frac{1}{N_c^d}\sum_{\substack{y=c\\({\bm x},y)\in\mathcal{D}_d}}{\bm x},
\end{equation}
where $N_c^d$ is the number of images with class $c$ in domain $d$. The domain distance $\texttt{dist}(\cdot, \cdot)$ and averaged class distance between domains $\texttt{class\_dist}(\cdot, \cdot)$ are defined as follows. Since different models have a feature space with a different dimensions and different metrics, we normalize the distance of two centroids by in-domain and in-class distance, respectively:
\begin{align}
    \texttt{cosine\_dist}({\bm x}_i, {\bm x}_j) & =1-\frac{{\bm x}_i\cdot {\bm x}_j}{\|{\bm x}_i\|\cdot\|{\bm x}_j\|}\\
    \texttt{in\_dist}(D_i) &= \frac{1}{|D_i|}\sum_{{\bm x}\in D_i}\texttt{cosine\_dist}({\bm x},\texttt{cent}_i)\\
    \texttt{dist}(D_i, D_j) & =\frac{\texttt{cosine\_dist}(\texttt{cent}_i, \texttt{cent}_j)}{\frac{1}{2}\big(\texttt{in\_dist}(D_i)+\texttt{in\_dist}(D_j)\big)}\\
    \texttt{class\_dist}(D_i, D_j) &= \frac{1}{n_c}\sum_{i=1}^{n_c}\texttt{dist}(D_i^c, D_j^c),
\end{align}
where $n_c$ is the number classes and $D_i^c$ is the set of images with class $c$ in domain $i$. 

\section{Additional Backbone Details}

The detailed backbone information is provided in Table~\ref{tab:backbone}.

\begin{table}[!ht]
\caption{Comparison between different backbones.}
\label{tab:backbone}
\begin{center}
\begin{tabular}{lcc}
\toprule
Model & \# Params & ImageNet Acc@1 (\%) \\
\midrule
AlexNet  & 61M & 56.5 \\
ResNet-18 & 11M & 69.8\\
ResNet-50 & 23M & 76.1\\
ViT-Base/16 & 86M & 81.1 \\
\bottomrule
\end{tabular}
\end{center}
\end{table}

\section{Ablation Study on Prompt Length}
\label{sec:ablpl}

We choose the prompt length to be 4 according to the experimental performance. We experiment with prompt length [2,4,8,16,32] and find that different prompt length does not greatly affect the performance of \ours~ under our experimental setting. However, we do notice that a bigger prompt length requires a higher learning rate to reach the best performance. To make the search space smaller and ensure a fair search space compared with the other methods as shown in Table~\ref{tab:full} in the appendix, we set the prompt length to a fixed length. The accuracy of different prompt lengths under our test protocol on OfficeHome is shown in Table~\ref{tab:ablpl}.

\begin{table}[!hbtp]
\caption{Accuracy (\%) on OfficeHome with different prompt length.}
\label{tab:ablpl}
\begin{center}
\begin{tabular}{lccccc}
\toprule
Length & 2 & 4 & 8 & 16 & 32 \\
\midrule
Acc (\%) & 75.5 & 76.0 & 75.9 & 76.0 & 75.8 \\
\bottomrule
\end{tabular}
\end{center}
\end{table}

\section{Additional Implementation Details and Hyper-parameter Selection}
\label{sec:hp}

\begin{table}[!h]
\centering
\caption{Hyper-parameter search space comparison.}
\resizebox{\textwidth}{!}{\small
\begin{tabular}{cccc}
\hline
\textbf{Hyper-parameter}  & \multicolumn{2}{c}{\textbf{Random Search}} & \textbf{Grid Search} \\ 
& DomainBed~\cite{gulrajani_search_2020} & Ensemble~\cite{arpit_ensemble_2021} & Ours \\
\hline
Learning rate (LR) & $10^{\text{Uniform}(-5, -3.5)}$ & $5e^{-5}$ & $\{1,5\}\times 10^{\{-6,-5,-4\}}$\\
Batch size & $2^{\text{Uniform}(3, 5.5)}$ & 32 & 32\\
Classifier LR multipler & - & - & $\{1, 10, 100\}$ \\
Last layer dropout~\cite{Huang2016DeepNW} & $\{0, 0.1, 0.5\}$ & $\{0, 0.1, 0.5\}$ & $\{0, 0.1, 0.5\}$ \\
Weight decay & $10^{\text{Uniform}(-6, -2)}$ & $10^{\text{Uniform}(-6, -4)}$ & 10$^{\{-2, -3, -4, -5\}}$ \\
\hline
\end{tabular}
}
\label{tab:hyper_param}
\end{table}

\begin{table}[!ht]
\centering
\caption{Hyper-parameter search space for different algorithms. Recommended one are in bold text.}
\begin{tabular}{lccc}
\toprule
\textbf{Method} & \textbf{Hyper-parameter}  & \textbf{Random Search} & \textbf{Grid Search} \\ 
& & DomainBed~\cite{gulrajani_search_2020} &  Ours \\
\hline
\makecell{MMD \\ CORAL} & $\gamma$ & $10^{\text{Uniform}(-1, 1)}$ & $\{\textbf{0.1}, 1, 10\}$\\
\hline
\multirow{5}{*}{\makecell{DANN \\ CDANN}} & $\gamma$ & $10^{\text{Uniform}(-2, 2)}$ & $\{\textbf{0.1}, 1, 10\}$ \\
 & discriminator weight decay & $10^{\text{Uniform}(-6, -2)}$ & same to weight decay\\
 & discriminator steps & $2^{\text{Uniform}(0,3)}$ & no need\\
 & gradient penalty & $10^{\text{Uniform}(-2, 1)}$ & no need \\
 & adam $\beta_1$ & $\{0, 0.5\}$ & 0.5 \\
\hline
MLDG & $\beta$ & $10^{\text{Uniform}(-1, 1)}$ & $\{0.1, 1, \textbf{10}\}$ \\ \hline
GroupDRO & $\eta$ & $10^{\text{Uniform}(-1, 1)}$ & $\{\textbf{0.01}, 0.1, 1, 10\}$ \\ \hline
\multirow{2}{*}{IRM} & $\lambda$ & $10^{\text{Uniform}(-1, 5)}$ & $\{\textbf{0.1},1,10,100,1000\}$ \\ 
 & iterations of penalty annealing & $10^{\text{Uniform}(0, 4)}$ & 500 \\ \hline
Mixup & $\alpha$ & 0.1 & $\{0.1, \textbf{0.2}, 0.5\}$ \\
\hline
\ours & $\lambda$ & -- & \{0.1, \textbf{1}, 10\} \\
\bottomrule
\end{tabular}
\label{tab:hyper_param2}
\end{table}

For domain alignment-based methods, we apply the alignment loss on the class token at the last layer. The overall loss is $\mathcal{L}=\mathcal{L}_{\text{ERM}} + \gamma\mathcal{L}_{\text{Align}}$ where the align loss is different for MMD, CORAL, DANN and CDANN. The discriminator-based methods DANN and CDANN implemented in~\cite{gulrajani_search_2020} did not follow the original one proposed in~\cite{ganin_domain-adversarial_2016}. We implement the gradient reverse layer, use global weight decay for the discriminator, and adopt $\beta_1=0.5$ for Adam optimizer as suggested in~\cite{radford2015unsupervised}. In this way, we reduce the search space needed for discriminator-based methods. For MLDG, following~\cite{gulrajani_search_2020}, we adopt the first-order formulation~\cite{Finn2017ModelAgnosticMF} for a fair comparison with other first-order methods with respect to computing budget. For GroupDRO, IRM, and Mixup, we sample several hyper-parameters in the random search space in~\cite{gulrajani_search_2020} at a different scale. The hyper-parameters for different methods in the second phase are listed in Table~\ref{tab:hyper_param2}.

\section{Additional Dataset Details}

The detailed dataset information is provided in Table \ref{tab:dataset}.

\begin{table}[!ht]
\centering
\caption{Detailed dataset statistics}
\label{tab:dataset}
\begin{tabular}{ccccc}
\toprule[1.3pt]
Dataset & \multicolumn{1}{c}{Domain} & \# image & \# image & \# class \\ \midrule
\multirow{4}{*}{PACS~\cite{Li2017DeeperBA}} & \ulbf{A}rt & 2,048 & \multirow{4}{*}{9,991} & \multirow{4}{*}{7} \\ \cline{2-3}
 & \ulbf{P}hoto & 1,670 & & \\ \cline{2-3}
 & \ulbf{C}lipart & 2,344 & & \\ \cline{2-3}
 & \ulbf{S}ketch & 3,929 & & \\ \midrule
 
\multirow{4}{*}{VLCS~\cite{Fang2013UnbiasedML}} & \multicolumn{1}{c}{\ulbf{V}OC2007} & 3,376 & \multirow{4}{*}{10,729} & \multirow{4}{*}{5} \\ \cline{2-3}
 & \multicolumn{1}{c}{\ulbf{L}abelMe} & 2,656 & & \\ \cline{2-3}
 & \multicolumn{1}{c}{\ulbf{S}UN09} & 3,282 & & \\ \cline{2-3}
 & \multicolumn{1}{c}{\ulbf{C}altech101} & 1,415 & & \\ \midrule
 
\multirow{4}{*}{OfficeHome~\cite{Venkateswara2017DeepHN}} & \multicolumn{1}{c}{\ulbf{A}rt} & 2,427 & \multirow{4}{*}{30,475} & \multirow{4}{*}{65} \\ \cline{2-3}
 & \multicolumn{1}{c}{\ulbf{C}lipart} & 4,365  & & \\ \cline{2-3}
 & \multicolumn{1}{c}{\ulbf{P}roduct} & 4,439  & & \\ \cline{2-3}
 & \multicolumn{1}{c}{\ulbf{R}eal} & 4,357 & & \\ \midrule
 
\multirow{6}{*}{DomainNet~\cite{Peng2019MomentMF}} 
& \multicolumn{1}{c}{\ulbf{C}lipart} & 48,129 & \multirow{6}{*}{586,575} & \multirow{6}{*}{345} \\ \cline{2-3}
& \multicolumn{1}{c}{\ulbf{I}nfograph} & 51,605 & & \\ \cline{2-3}
& \multicolumn{1}{c}{\ulbf{P}ainting} & 72,266 & & \\ \cline{2-3}
& \multicolumn{1}{c}{\ulbf{Q}uickdraw} & 172,500 & & \\ \cline{2-3}
& \multicolumn{1}{c}{\ulbf{R}eal} & 172,947 & & \\ \cline{2-3}
& \multicolumn{1}{c}{\ulbf{S}ketch} & 69,128 & & \\ \midrule

\end{tabular}
\end{table}

\end{document}